\documentclass[letterpaper]{article}
\usepackage{aaai2027}

\usepackage[hyphens]{url}
\usepackage{graphicx}
\usepackage{natbib}
\usepackage{caption}
\usepackage{amsmath}
\usepackage{amssymb}
\usepackage{booktabs}
\usepackage{multirow}

\title{TALON: A Temporally Aware Longitudinal Framework for Radiology Report Generation}

\author{
Nien-Tsyr Sun$^{1}$,
Min-Chen Chen$^{2}$,
Huei-Nian Hong$^{3}$,
Vincent S. Tseng$^{4}$
}

\affiliations{
Institute of Data Science and Engineering, National Yang Ming Chiao Tung University, Hsinchu 300093, Taiwan$^{1}$\\
Institute of Computer Science and Engineering, National Yang Ming Chiao Tung University, Hsinchu 300093, Taiwan$^{2}$\\
Institute of Statistics, National Yang Ming Chiao Tung University, Hsinchu 300093, Taiwan$^{3}$\\
Department of Computer Science, National Yang Ming Chiao Tung University, Hsinchu 300093, Taiwan$^{4}$\\[0.6em]

sallyball838.cs13@nycu.edu.tw$^{1}$\\
natalie1116.cs11@nycu.edu.tw$^{2}$\\
hhung@stat.nycu.edu.tw$^{3}$\\
vtseng@cs.nycu.edu.tw$^{4}$
}

\begin{document}
\maketitle

\begin{abstract}
Current radiology report generation (RRG) models usually produce descriptive reports based on a single examination or only the most recent prior examination, limiting their ability to perform accurate and meaningful longitudinal comparisons and detect subtle interval changes. Although recent approaches have begun to incorporate multiple prior examinations, they usually aggregate a fixed-length history without explicitly modeling the role-dependent relevance of each prior examination before fusion. To address this, we propose TALON, a \textbf{T}emporally \textbf{A}ware \textbf{LON}gitudinal RRG framework that adaptively integrates variable-length patient histories. The underlying Dual-Channel Temporal Fusion Module (DCTFM) compares the current examination with each prior examination through complementary similarity and change channels to capture persistent findings and interval changes, respectively. The specially designed channel-specific attention estimates the relevance of each prior examination, while a learned prior-specific gate adaptively integrates informative longitudinal evidence and suppresses redundancy. Experiments on MIMIC-CXR show that TALON outperforms the current state-of-the-art method on various clinical efficacy and graph-based metrics. When more prior examinations become available, TALON's performance on these metrics improves even further, emphasizing the strength of TALON's DCTFM in modeling longitudinal RRG across longer and more complex patient histories than existing approaches.
\end{abstract}

\section{Introduction}

Radiology report generation (RRG) translates medical images into findings that can support radiologist review~\cite{chen2020r2gen,sloan2024review}. Most systems nevertheless treat an examination as an isolated image--text pair~\cite{chen2021cmn,jin2024promptmrg,liu2024bllm}. This formulation is poorly matched to follow-up radiology, where the current chest X-ray is compared with prior examinations to determine what is new, resolving, progressing, or unchanged. Without that reference, a model may omit subtle interval change, overcall a chronic abnormality as acute, or hallucinate a comparison to a nonexistent prior~\cite{ramesh2022hallucinated}.

Longitudinal context is valuable, but more history is not automatically better history. Temporal-gap modeling, redundancy mitigation, and cross-time spatial alignment remain open problems in longitudinal radiology report generation (LRRG)~\cite{zhou2025longitudinalreview}. Priors differ in recency, projection, quality, and clinical relevance: a nearby study may expose acute change, whereas an older study may establish chronicity. Indiscriminate pooling can therefore dilute interval evidence or amplify redundant patterns.

Our research starts from a simple abstraction of the radiologist's comparison process. Longitudinal interpretation is not a single history-retrieval operation, but two complementary evidence searches: \emph{stability evidence} establishes what persists across visits, whereas \emph{change evidence} identifies what is new or insufficiently explained by prior examinations. Crucially, the relevance of a prior is conclusion-dependent. A nearly identical prior can be decisive for establishing chronicity yet uninformative for localizing a new finding; conversely, a discrepant prior can expose interval change without serving as a reliable stability reference. A single temporal attention distribution forces these opposing roles into one ranking.

Existing LRRG methods introduce temporal constraints, patient-specific context, and multi-history input~\cite{liu2025hcllm,liu2025mlrg,liu2026priorrg,wang2024hergen,yang2025stream}. However, paired systems expose only a recent prior, while multi-history systems encode or fuse the available examinations without explicitly estimating the relevance of each prior before temporal aggregation. The unresolved problem is therefore not simply accepting more history, but performing \emph{prior-wise selection within history} while preserving which examination supports each temporal conclusion.

To address this problem, we propose TALON, a longitudinal framework centered on a Dual-Channel Temporal Fusion Module (DCTFM). Each prior is aligned to the current study but remains an explicit comparison candidate. Separate similarity and change channels then estimate how strongly that prior supports persistent structure and interval evidence. A prior-specific gate routes the selected evidence through co-activation and current-minus-prior residual streams, while zero-initialized gains preserve a direct current-image path when history is absent or unhelpful. The key technical contribution is this \emph{prior-wise, criterion-specific selection before aggregation}: the same examination may be useful for establishing stability but less informative for identifying change, or vice versa.

The main contributions of this work are summarized as follows:
\begin{itemize}
    \item We formulate variable-length longitudinal report generation as explicit current--prior comparison, preserving prior identity across first visits, single-prior cases, and deeper patient histories within one computation graph.
    \item We introduce the Dual-Channel Temporal Fusion Module, which learns separate prior relevance distributions for stability and interval change and transforms them through gated co-activation and residual streams before aggregation.
    \item We validate the proposed design on MIMIC-CXR. TALON leads the clinical efficacy metrics and RadGraph-F1 while ranking first or second on every language-generation metric. Controlled analyses further show larger gains when deeper history is available.
\end{itemize}

\section{Related Work}

\subsection{Radiology Report Generation}

Early RRG systems commonly used encoder--decoder architectures with recurrent or Transformer decoders~\cite{chen2020r2gen}. Later work improved cross-modal alignment through memory networks~\cite{chen2021cmn,shen2024man}, knowledge-aware representations~\cite{huang2023kiut}, and contrastive objectives~\cite{li2024cofe}. Recent large-language-model approaches focus on adapting general text generation to radiology. B-LLM uses in-domain induction and coarse-to-fine decoding~\cite{liu2024bllm}, PromptMRG converts diagnostic predictions into generation prompts~\cite{jin2024promptmrg}, and MPO optimizes report generation for multiple quality preferences~\cite{xiao2025mpo}. These methods improve linguistic or clinical quality, but their core formulation remains centered on the current examination.

Medical vision--language pre-training supplies domain-specific text and image representations~\cite{boecking2022cxrbert,perezgarcia2024raddino,bannur2023biovilt}. TALON uses these encoders but contributes the temporal operator between encoding and generation, not a new foundation model.

\subsection{Longitudinal Radiology Report Generation}

Longitudinal RRG introduces prior images, reports, or both to model disease evolution~\cite{zhou2025longitudinalreview}. Paired-visit methods use current--prior representations, report prefilling, shared/specific constraints, or difference-aware residuals~\cite{serra2023controllable,ramesh2022hallucinated,liu2025hcllm,song2025ddatr}. MLRG combines visual history and patient context in two-stage training~\cite{liu2025mlrg}; PriorRG uses context-guided pre-training and coarse-to-fine decoding~\cite{liu2026priorrg}.

Multi-history methods move beyond paired visits: HERGen organizes longitudinal visit groups, while STREAM encodes current and historical studies as visual prompts~\cite{wang2024hergen,yang2025stream}. These approaches demonstrate the value of deeper context, but their principal formulations do not explicitly retain each examination as a scored candidate for relevance screening before temporal aggregation. This distinction matters because complete histories introduce redundancy, and unregistered cross-time patches can confound progression~\cite{zhou2025longitudinalreview}. TALON therefore preserves prior identity through current-conditioned alignment and estimates the contribution of every current--prior comparison before aggregation.

Relative to the two-stage systems of MLRG and PriorRG~\cite{liu2025mlrg,liu2026priorrg}, TALON addresses a different question: how should variable-length histories be screened when individual priors differ in relevance? DCTFM retains indexed priors until their relevance has been estimated and only then aggregates the selected evidence. Its similarity and change channels serve as complementary scoring criteria within this prior-wise screening operation, rather than constituting the cross-method distinction by themselves.

\section{Method}
\label{sec:method}

\begin{figure*}[t]
\centering
\includegraphics[width=\textwidth]{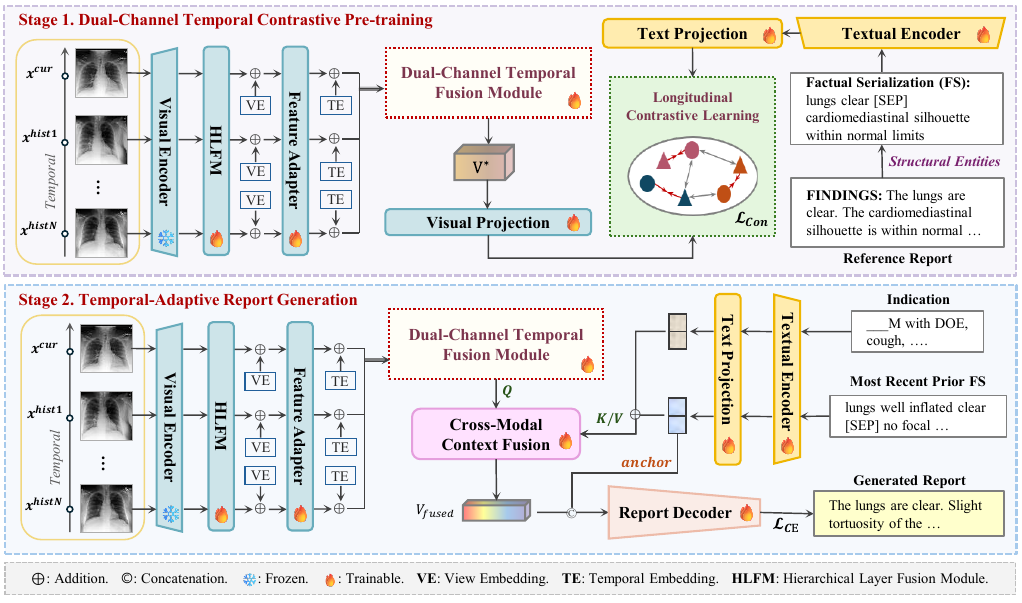}
\caption{Overall TALON architecture. Stage~1 performs Longitudinal Contrastive Learning by aligning the history-enhanced visual representation with the factual serialization (FS) of the current Findings; Stage~2 transfers the visual-temporal modules for report generation. HLFM, VE, TE, DCTFM, and CMCF denote Hierarchical Layer Fusion Module, view embedding, temporal embedding, Dual-Channel Temporal Fusion Module, and Cross-Modal Context Fusion.}
\label{fig:framework}
\end{figure*}

\subsection{Problem Formulation}

For the $i$-th current examination, let $x_i^0$ denote the current chest X-ray and
\begin{equation}
\mathcal{H}_i=\{(x_i^n,\Delta t_i^n)\}_{n=1}^{N_i}, \qquad 0\leq N_i\leq N_{\max},
\end{equation}
denote its reverse-chronological priors, where $n=1$ is the most recent and $\Delta t_i^n$ is the elapsed time to the current study. The model also receives the indication $z_i$ and the factual serialization (FS) $r_i^1$ of the most recent prior report when available. TALON learns $p_\theta(y_i\mid x_i^0,\mathcal{H}_i,z_i,r_i^1)$ for current Findings $y_i$, covering first visits, single-prior, and multi-history cases in one graph.

As shown in Figure~\ref{fig:framework}, TALON follows the two-stage contrastive-to-generative paradigm of MLRG and PriorRG~\cite{liu2025mlrg,liu2026priorrg} while sharing one prior-wise temporal operator across both stages. Each prior is aligned to the current study, scored separately for stability and change, transformed in two gated streams, and then aggregated. Stage~1 aligns the resulting visual representation with current-report FS; Stage~2 transfers the visual-temporal modules for generation with patient-specific text.

\subsection{Hierarchical Visual and Temporal Encoding}

A frozen RAD-DINO encoder~\cite{perezgarcia2024raddino} extracts image tokens. Inspired by attention-enhanced multi-layer fusion and CBAM-style recalibration~\cite{liu2026priorrg,woo2018cbam}, the Hierarchical Layer Fusion Module (HLFM) combines layers $6$, $9$, and $12$ with patch-wise depth weights:
\begin{align}
w_{p}^{\ell}&=\operatorname{softmax}_{\ell}\!\left(\operatorname{HLFM}(\{H_{p}^{6},H_{p}^{9},H_{p}^{12}\})\right),\nonumber\\
V_p&=\operatorname{Adapter}\!\left(\sum_{\ell\in\{6,9,12\}}w_p^{\ell}H_p^{\ell}\right).
\label{eq:hlfm}
\end{align}
Each study receives three learned codes. View embedding (VE) identifies the acquisition projection, history-position embedding distinguishes the current study from the $n$-th most recent prior, and temporal embedding (TE) represents the elapsed days. TE uses 13 increasingly coarse indices, from same-day intervals to gaps longer than five years, so chronological rank and absolute temporal distance remain distinguishable.

\subsection{Dual-Channel Temporal Fusion Module}
\label{sec:dctfm}

\begin{figure}[t]
\centering
\includegraphics[width=\columnwidth]{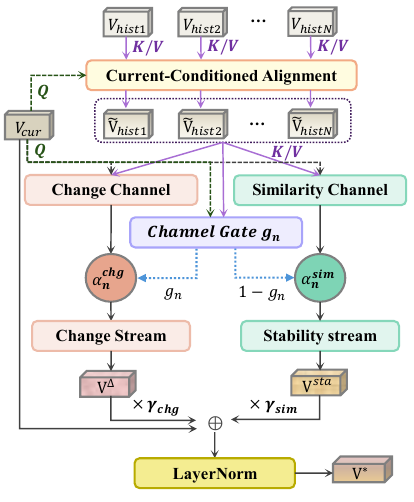}
\caption{Dual-Channel Temporal Fusion Module (DCTFM). Aligned priors are independently weighted by similarity and change channels, gated into stability and residual-change streams, and integrated with the current representation.}
\label{fig:dctfm}
\end{figure}

Let $V_0\in\mathbb{R}^{L\times d}$ and $V_n$ denote current and prior-$n$ tokens, corresponding to $V_{\mathrm{cur}}$ and $V_{\mathrm{hist}n}$ in Figure~\ref{fig:dctfm}.

\paragraph{Current-Conditioned Spatial Alignment.}
Cross-visit projection and positioning differences make direct patch subtraction unreliable. TALON first aligns each prior to the current representation:
\begin{equation}
\widetilde{V}_n=\operatorname{MHA}(Q=V_0,K=V_n,V=V_n).
\label{eq:align}
\end{equation}
$V_0$ remains the query for every prior, preserving a current-conditioned comparison path without collapsing prior identity.

\paragraph{Similarity Channel.}
Let $q=V_0[\mathrm{CLS}]$ and $h_n=\widetilde{V}_n[\mathrm{CLS}]$, and let $(\cdot)_h$ denote the $h$-th head slice. Using an image-only query, the similarity channel scores priors that support persistent structure:
\begin{align}
\bar q^{\mathrm{sim}}&=W_q^s q,\qquad \bar h_n^{\mathrm{sim}}=W_k^s h_n,\nonumber\\
e_{nh}^{\mathrm{sim}}&=\frac{s\,(\bar q^{\mathrm{sim}})_h^{\top}(\bar h_n^{\mathrm{sim}})_h}{\sqrt{d_h}},\nonumber\\
\alpha_n^{\mathrm{sim}}&=\frac{1}{K}\sum_{h=1}^{K}\operatorname{softmax}_{n}(e_{nh}^{\mathrm{sim}}),
\label{eq:sim}
\end{align}
where $s$ is learned and the softmax is restricted to valid priors. Values are not mixed at this stage, so prior identity is retained.

\paragraph{Change Channel.}
To score interval evidence independently of visual persistence, the change channel uses separate query and key projections:
\begin{align}
\bar q^{\mathrm{chg}}&=\operatorname{LN}(W_q^c q),\qquad \bar h_n^{\mathrm{chg}}=W_k^c h_n,\nonumber\\
e_{nh}^{\mathrm{chg}}&=\frac{s\,(\bar q^{\mathrm{chg}})_h^{\top}(\bar h_n^{\mathrm{chg}})_h}{\sqrt{d_h}},\nonumber\\
\alpha_n^{\mathrm{chg}}&=\frac{1}{K}\sum_{h=1}^{K}\operatorname{softmax}_{n}(e_{nh}^{\mathrm{chg}}),
\label{eq:change}
\end{align}
The reported configuration injects no indication into DCTFM. Because both scores are normalized across valid priors, additional history creates competition within each role rather than uniformly increasing historical evidence.

\paragraph{Channel Gate.}
DCTFM predicts a prior-specific balance between change and stability:
\begin{equation}
g_n=\sigma\!\left(\operatorname{MLP}(\operatorname{LN}([\bar q^{\mathrm{chg}};h_n]))\right).
\label{eq:gate}
\end{equation}
Its zero-initialized final layer gives $g_n=0.5$ initially, avoiding a preset channel preference. The feature computation retains both relevance distributions rather than merging them into one attention map.

\paragraph{Factorized Temporal Streams.}
For each prior, a second cross-attention estimates the portion of the current sequence supported by the aligned prior:
\begin{equation}
R_n=\operatorname{LN}(\operatorname{MHA}(V_0,\widetilde{V}_n,\widetilde{V}_n)),\qquad C_n=V_0-R_n.
\label{eq:residual}
\end{equation}
$R_n$ is the prior-explainable component, and the signed residual $C_n$ retains current evidence not reconstructed from prior $n$. The change stream transforms $[V_0;C_n]$. The stability stream instead uses current tokens, aligned prior tokens, and their element-wise co-activation:
\begin{align}
V^{\Delta}&=\sum_{n=1}^{N}\alpha_n^{\mathrm{chg}}g_n\,\phi_{\Delta}([V_0;C_n]),\nonumber\\
V^{\mathrm{sta}}&=\sum_{n=1}^{N}\alpha_n^{\mathrm{sim}}(1-g_n)\,\phi_{S}([V_0;\widetilde{V}_n;V_0\odot\widetilde{V}_n]).
\label{eq:streams}
\end{align}
The residual stream emphasizes unexplained current evidence, whereas co-activation preserves structures supported across visits. Both transformations occur before aggregation, preventing early cancellation between the two evidence types.

\paragraph{Residual Integration.}
The final visual representation is
\begin{equation}
V^{*}=\operatorname{LN}(V_0+\gamma_{\mathrm{chg}}V^{\Delta}+\gamma_{\mathrm{sim}}V^{\mathrm{sta}}),
\label{eq:fusion}
\end{equation}
Both gains are zero-initialized, so training begins from the current-image path. Validity masking sets both stream sums to zero when history is unavailable, reducing the model to the normalized current representation.

\subsection{Cross-Modal Context Fusion and Generation}

Following MLRG's patient-context integration~\cite{liu2025mlrg}, CXR-BERT~\cite{boecking2022cxrbert} encodes the indication and most recent prior-report FS with segment identifiers, and a projection maps the text tokens to dimension $d$. CMCF uses $V^{*}$ as query and the projected text as key/value. The projected prior-FS tokens are additionally appended as a late language anchor before warm-started DistilGPT2 decoding~\cite{nicolson2023warm}. TALON therefore selects from up to five prior images while using only the most recent report FS as textual history.

\begin{table*}[t]
\centering
\setlength{\tabcolsep}{3.3pt}
\renewcommand{\arraystretch}{1.3}
\begin{tabular}{@{} c|ccc|cccc|cccccc @{}}
\hline
\multirow{2}{*}{\textbf{Dataset}} & \multirow{2}{*}{\textbf{Model}} & \multirow{2}{*}{\textbf{Venue}} & \multicolumn{1}{c|}{\multirow{2}{*}{\textbf{History}}} & \multicolumn{4}{c|}{\textbf{CE Metrics}} & \multicolumn{6}{c}{\textbf{NLG Metrics}} \\
\cline{5-8}\cline{9-14}
& & & & \textbf{P} & \textbf{R} & \textbf{F1} & \textbf{RG} & \textbf{B-1} & \textbf{B-2} & \textbf{B-3} & \textbf{B-4} & \textbf{MTR} & \textbf{R-L} \\
\hline
\multirow{12}{*}{MIMIC-CXR}
& CoFE    & ECCV'24   & $S$     & 0.489 & 0.370 & 0.405 & --    & --             & --             & --             & 0.125 & 0.176 & 0.304 \\
& DCG     & ACMMM'24  & $S$     & 0.441 & 0.414 & 0.404 & --    & 0.397          & 0.258          & 0.166          & 0.126 & 0.162 & 0.295 \\
& MAN     & AAAI'24   & $S$     & 0.411 & 0.398 & 0.389 & --    & 0.396          & 0.244          & 0.162          & 0.115 & 0.151 & 0.274 \\
& R2-LLM  & AAAI'24   & $S$     & 0.465 & 0.482 & 0.473 & --    & 0.402          & 0.262          & 0.180          & 0.128 & 0.175 & 0.291 \\
& SEI     & MICCAI'24 & $S$     & 0.523 & 0.410 & 0.460 & 0.249 & 0.382          & 0.247          & 0.177          & 0.135 & 0.158 & 0.299 \\
& HERGen  & ECCV'24   & $L_N$   & --    & --    & --    & --    & 0.395          & 0.248          & 0.169          & 0.122 & 0.156 & 0.285 \\
& Med-LMM & ACMMM'24  & $S$     & 0.412 & 0.373 & 0.395 & --    & --             & --             & --             & 0.128 & 0.161 & 0.289 \\
& MPO     & AAAI'25   & $S$     & 0.436 & 0.376 & 0.353 & --    & 0.416          & 0.269          & 0.191          & 0.139 & 0.162 & 0.309 \\
& STREAM  & TMI'25    & $L_N$   & 0.515 & 0.447 & 0.478 & 0.223 & \textbf{0.437} & 0.278          & 0.192          & 0.139 & 0.172 & 0.297 \\
& MLRG    & CVPR'25   & $L_1$   & \underline{0.549} & 0.468 & 0.505 & 0.291 & 0.411 & 0.277 & 0.204 & 0.158 & 0.176 & 0.320 \\
& PriorRG & AAAI'26   & $L_1$   & 0.541 & \underline{0.485} & \underline{0.511} & \underline{0.296} & 0.412 & \underline{0.290} & \textbf{0.220} & \textbf{0.175} & \textbf{0.189} & \underline{0.324} \\
\cline{2-14}
& \textbf{TALON} & -- & $L_N$ & \textbf{0.552} & \textbf{0.501} & \textbf{0.525} & \textbf{0.305} & \underline{0.431} & \textbf{0.293} & \underline{0.217} & \underline{0.169} & \underline{0.186} & \textbf{0.325} \\
\hline
\end{tabular}
\caption{Comparison with state-of-the-art methods on MIMIC-CXR using clinical efficacy (CE) and natural language generation (NLG) metrics. History settings $S$, $L_1$, and $L_N$ denote current-only, one-prior, and multi-prior inputs. Best and second-best results are \textbf{bolded} and \underline{underlined}, respectively.}
\label{tab:main}
\end{table*}

\begin{table*}[t]
\centering
\setlength{\tabcolsep}{1.2pt}
\renewcommand{\arraystretch}{1.3}
\begin{tabular*}{\textwidth}{@{\extracolsep{\fill}}ccccc cccc cccccc}
\toprule
\multirow{2}{*}{\textbf{PT}} & \multirow{2}{*}{\textbf{Sim}} & \multirow{2}{*}{\textbf{Chg}} & \multirow{2}{*}{\textbf{Anc}} & \multirow{2}{*}{\textbf{CMCF}} & \multicolumn{4}{c}{\textbf{CE Metrics}} & \multicolumn{6}{c}{\textbf{NLG Metrics}} \\
\cmidrule(lr){6-9}\cmidrule(lr){10-15}
& & & & & \textbf{P} & \textbf{R} & \textbf{F1} & \textbf{RG} & \textbf{B-1} & \textbf{B-2} & \textbf{B-3} & \textbf{B-4} & \textbf{MTR} & \textbf{R-L} \\
\midrule
\checkmark & -- & -- & \checkmark & \checkmark & 0.543 & 0.488 & 0.514 & 0.292 & 0.428 & 0.289 & 0.213 & 0.166 & 0.184 & 0.321 \\
\checkmark & \checkmark & -- & \checkmark & \checkmark & 0.548 & 0.492 & 0.519 & 0.305 & 0.428 & 0.291 & 0.216 & 0.168 & 0.185 & \textbf{0.325} \\
\checkmark & -- & \checkmark & \checkmark & \checkmark & 0.547 & 0.500 & 0.522 & \textbf{0.307} & \textbf{0.431} & 0.292 & 0.216 & 0.168 & \textbf{0.186} & 0.324 \\
-- & \checkmark & \checkmark & \checkmark & \checkmark & 0.550 & 0.488 & 0.517 & 0.299 & 0.430 & 0.291 & 0.214 & 0.167 & \textbf{0.186} & 0.323 \\
\checkmark & \checkmark & \checkmark & -- & \checkmark & 0.547 & 0.481 & 0.512 & 0.292 & 0.406 & 0.271 & 0.197 & 0.152 & 0.176 & 0.315 \\
\checkmark & \checkmark & \checkmark & \checkmark & -- & 0.536 & \textbf{0.501} & 0.518 & 0.291 & 0.422 & 0.280 & 0.202 & 0.155 & 0.180 & 0.312 \\
\midrule
\checkmark & \checkmark & \checkmark & \checkmark & \checkmark & \textbf{0.552} & \textbf{0.501} & \textbf{0.525} & 0.305 & \textbf{0.431} & \textbf{0.293} & \textbf{0.217} & \textbf{0.169} & \textbf{0.186} & \textbf{0.325} \\
\bottomrule
\end{tabular*}
\caption{Ablation study on MIMIC-CXR. PT, Sim, Chg, Anc, and CMCF denote contrastive pre-training, similarity/stability channel, change channel, prior-report FS anchor, and Cross-Modal Context Fusion. The first row disables both DCTFM streams but retains the textual history path. The final row is TALON; unchecked entries indicate removed components. Metric abbreviations follow Table~\ref{tab:main}; best results are bolded.}
\label{tab:ablation}
\end{table*}

\begin{table}[t]
\centering
\small
\setlength{\tabcolsep}{1pt}
\renewcommand{\arraystretch}{1.2}
\begin{tabular*}{\columnwidth}{@{\extracolsep{\fill}}l cccc cc}
\toprule
\multirow{2}{*}{\textbf{Visible priors}} & \multicolumn{4}{c}{\textbf{CE Metrics}} & \multicolumn{2}{c}{\textbf{NLG Metrics}} \\
\cmidrule(lr){2-5}\cmidrule(lr){6-7}
& \textbf{P} & \textbf{R} & \textbf{F1} & \textbf{RG} & \textbf{B-1} & \textbf{R-L} \\
\midrule
Most recent 1 & \textbf{0.573} & 0.525 & 0.548 & 0.301 & 0.433 & \textbf{0.322} \\
Most recent 3 & 0.571 & 0.530 & 0.549 & \textbf{0.302} & 0.433 & \textbf{0.322} \\
Most recent 5 & \textbf{0.573} & \textbf{0.534} & \textbf{0.553} & \textbf{0.302} & \textbf{0.434} & \textbf{0.322} \\
\bottomrule
\end{tabular*}
\caption{Fixed-cohort history masking on the same 1,920 test samples with at least five valid priors. One TALON checkpoint is evaluated while exposing only the most recent 1, 3, or 5 priors.}
\label{tab:history_masking}
\end{table}

\begin{table}[t]
\centering
\small
\setlength{\tabcolsep}{1pt}
\renewcommand{\arraystretch}{1.2}
\begin{tabular*}{\columnwidth}{@{\extracolsep{\fill}}l cccc cc}
\toprule
\multirow{2}{*}{\textbf{Assignment}} & \multicolumn{4}{c}{\textbf{CE Metrics}} & \multicolumn{2}{c}{\textbf{NLG Metrics}} \\
\cmidrule(lr){2-5}\cmidrule(lr){6-7}
& \textbf{P} & \textbf{R} & \textbf{F1} & \textbf{RG} & \textbf{B-1} & \textbf{R-L} \\
\midrule
Uniform & 0.551 & 0.497 & 0.523 & \textbf{0.305} & \textbf{0.431} & 0.324 \\
Permuted & \textbf{0.552} & 0.498 & 0.524 & 0.303 & 0.429 & 0.322 \\
Learned & \textbf{0.552} & \textbf{0.501} & \textbf{0.525} & \textbf{0.305} & \textbf{0.431} & \textbf{0.325} \\
\bottomrule
\end{tabular*}
\caption{Prior-weight intervention on the full test set using the same TALON checkpoint.}
\label{tab:weight_intervention}
\end{table}

\begin{table}[t]
\centering
\small
\setlength{\tabcolsep}{0pt}
\renewcommand{\arraystretch}{1.2}
\begin{tabular*}{\columnwidth}{@{\extracolsep{\fill}}l ccc ccc c}
\toprule
\multirow{2}{*}{\textbf{Subset}} & \multicolumn{3}{c}{\textbf{PriorRG}} & \multicolumn{3}{c}{\textbf{TALON}} & \multirow{2}{*}{\textbf{$\Delta$F1}} \\
\cmidrule(lr){2-4}\cmidrule(lr){5-7}
& \textbf{P} & \textbf{R} & \textbf{F1} & \textbf{P} & \textbf{R} & \textbf{F1} & \\
\midrule
Full & 0.541 & 0.485 & 0.511 & \textbf{0.552} & \textbf{0.501} & \textbf{0.525} & $+1.4\%$ \\
No prior & 0.491 & \textbf{0.441} & 0.465 & \textbf{0.513} & 0.434 & \textbf{0.470} & $+0.6\%$ \\
$\geq1$ prior & 0.549 & 0.491 & 0.518 & \textbf{0.559} & \textbf{0.512} & \textbf{0.535} & $+1.6\%$ \\
$\geq2$ priors & 0.548 & 0.490 & 0.517 & \textbf{0.564} & \textbf{0.518} & \textbf{0.540} & $+2.2\%$ \\
$\geq3$ priors & 0.553 & 0.493 & 0.521 & \textbf{0.566} & \textbf{0.521} & \textbf{0.543} & $+2.1\%$ \\
$\geq4$ priors & 0.557 & 0.492 & 0.523 & \textbf{0.569} & \textbf{0.525} & \textbf{0.546} & $+2.3\%$ \\
$\geq5$ priors & 0.561 & 0.494 & 0.525 & \textbf{0.575} & \textbf{0.534} & \textbf{0.553} & $\mathbf{+2.8\%}$ \\
\bottomrule
\end{tabular*}
\caption{CheXbert precision (P), recall (R), and F1 by naturally available history. $\Delta$F1 denotes the absolute percentage-point gain of TALON over PriorRG; better paired results are bolded.}
\label{tab:natural_depth}
\end{table}

\begin{figure*}[t]
\centering
\includegraphics[width=\textwidth]{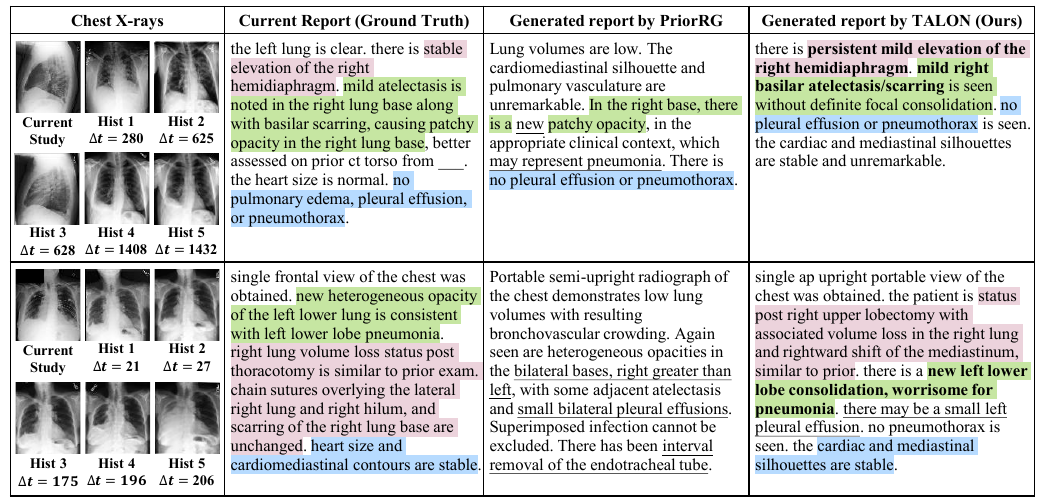}
\caption{Qualitative comparison between PriorRG and TALON on a
long-horizon stability case (top) and a short-interval change case
(bottom). Corresponding findings in the reference and generated
reports are highlighted in the same color. Key longitudinal
descriptions generated by TALON are marked in \textbf{bold}, while findings
inconsistent with or unsupported by the reference report are
\underline{underlined}. Hist~1--5 are ordered from the most recent to the oldest
prior study, and $\Delta t$ denotes the time interval in days.}
\label{fig:case}
\end{figure*}

\subsection{Training Objectives}

\paragraph{Stage 1: Longitudinal Contrastive Learning.}
Inspired by MLRG's structurally guided contrastive supervision~\cite{liu2025mlrg}, we use serialized structural entities extracted from the current Findings~\cite{liu2024sea} as text targets. CXR-BERT and visual/text projection heads map pooled $V^{*}$ and the target sequence into a shared space. For normalized embeddings $u_i,t_j$, let $s_{ij}=u_i^{\top}t_j/\tau$ and $p_{ij}$ indicate matching studies. We use the symmetric image--text contrastive objective~\cite{oord2018cpc}
\begin{equation}
\mathcal{L}_{\mathrm{con}}=-\frac{1}{2B}\sum_{i,j}p_{ij}\!\left[
\log\frac{e^{s_{ij}}}{\sum_k e^{s_{ik}}}+
\log\frac{e^{s_{ij}}}{\sum_k e^{s_{kj}}}\right].
\label{eq:contrastive}
\end{equation}
Stage~1 trains HLFM, temporal embeddings, DCTFM, both projection heads, and CXR-BERT while RAD-DINO remains frozen.

\paragraph{Stage 2: report generation.}
Stage~2 initializes the shared visual-temporal weights from Stage~1 and minimizes teacher-forced token cross-entropy:
\begin{equation}
\mathcal{L}_{\mathrm{CE}}=-\sum_{t=1}^{T}\log p_{\theta}(y_t\mid y_{<t},V^{*},z,r^1).
\label{eq:ce}
\end{equation}
Inference follows the same path with beam search.

\section{Experiments}
\label{sec:experiments}

\subsection{Experimental Settings}

\paragraph{Dataset.}
We evaluate on MIMIC-CXR~\cite{johnson2019mimiccxr}, a large-scale public chest X-ray dataset whose patient-linked studies, timestamps, and free-text reports support longitudinal reconstruction. Studies are ordered chronologically by patient; each sample uses one selected current-view image and report together with up to five nearest valid prior studies. Split composition and longitudinal statistics are reported in Supplementary Table~S1.

\paragraph{Evaluation metrics.}
We report BLEU-1--4~\cite{papineni2002bleu}, METEOR~\cite{banerjee2005meteor}, and ROUGE-L~\cite{lin2004rouge} for lexical generation quality. Clinical efficacy (CE) is measured by micro-averaged precision, recall, and F1 over 14 CheXbert observations~\cite{smit2020chexbert}; RadGraph-F1 evaluates entity--relation overlap~\cite{jain2021radgraph}. We treat CE and RadGraph as primary because radiology-specific evaluations show that clinically structured metrics align more closely with radiologist judgments, whereas lexical-overlap metrics can miss factual errors despite high surface similarity~\cite{yu2023evaluating,ostmeier2024green,song2025ddatr}. Because published NLG scores can also depend on text normalization and tokenization, cross-paper BLEU differences are interpreted descriptively; Supplementary Section~C provides further discussion.

\paragraph{Implementation details.}
We use $N_{\max}=5$, $d=768$, eight attention heads, and the RAD-DINO, CXR-BERT, and DistilGPT2 backbones described above. The five-prior cap balances historical coverage and computation; unavailable positions are masked, so TALON remains variable-length. Models are optimized with AdamW on NVIDIA H100 Tensor Core GPUs, selected by validation RadGraph-F1+BLEU-4+CheXbert micro-F1, and decoded with three beams and at most 100 new tokens. Additional optimization details are provided in Supplementary Section~A.

\subsection{Main Results}

We compare TALON with representative single-study and longitudinal RRG systems, including CoFE~\cite{li2024cofe}, DCG~\cite{liang2024dcg}, MAN~\cite{shen2024man}, B-LLM/R2-LLM~\cite{liu2024bllm}, SEI~\cite{liu2024sea}, HERGen~\cite{wang2024hergen}, Med-LMM~\cite{liu2024medlmm}, MPO~\cite{xiao2025mpo}, STREAM~\cite{yang2025stream}, MLRG~\cite{liu2025mlrg}, and PriorRG~\cite{liu2026priorrg}. Baseline values in Table~\ref{tab:main} are transcribed from their publications, with each method retaining its reported input setting, whereas TALON is evaluated with our pipeline. Supplementary Table~S8 summarizes the corresponding input taxonomy.

Table~\ref{tab:main} shows that TALON leads six of the ten reported metrics and ranks second on the remaining four. It achieves the strongest clinical efficacy and RadGraph-F1, with the most pronounced improvement in recall over the closest longitudinal baselines. The gains in CE precision, recall, F1, and RadGraph-F1 over MLRG and PriorRG are statistically significant at the $p\leq0.05$ level. Its advantage over multi-history STREAM indicates that access to multiple historical studies alone is insufficient. Meanwhile, competitive NLG performance shows that broader clinical coverage is achieved without a material loss of lexical similarity.

\paragraph{Observation-level clinical efficacy.}
Supplementary Table~S7 decomposes clinical efficacy over the 14 CheXbert observations. TALON exceeds PriorRG in F1 on 10 observations and leads six of the seven most prevalent observations. The aggregate improvement is therefore broad rather than driven by a single or predominantly rare category.

\subsection{Model Analysis}

\paragraph{Component ablations.}
Table~\ref{tab:ablation} isolates the contribution of each component. In the first row, both DCTFM stream gains are set to zero. Prior images are still encoded and aligned, but they are not fused into the final visual representation; the most recent prior-report FS anchor and CMCF remain active. Its lower clinical efficacy and RadGraph-F1 show that textual history alone does not explain TALON's gains. The single-channel variants recover complementary aspects: similarity-only preserves lexical fidelity, whereas change-only improves recall and achieves the highest RadGraph-F1. Combining both channels yields the highest CheXbert precision and F1 together with the best or tied-best NLG results, providing the strongest overall clinical--linguistic balance. Removing Stage~1 mainly reduces recall and F1. Removing the prior-report anchor causes the largest lexical decline, whereas bypassing CMCF weakens RadGraph-F1 and all NLG metrics. These controls separate the contributions of visual temporal fusion, contrastive initialization, and textual-context integration.

\paragraph{Prior-anchor repetition.}
The anchor ablation shows that prior FS supports language generation. Supplementary Table~S5 further shows that TALON's prior-text overlap and four-gram copy rate remain below the natural overlap between reference and prior reports, indicating that the gain is not explained by excessive copying.

\paragraph{Controlled history utilization.}
Tables~\ref{tab:history_masking} and~\ref{tab:weight_intervention} test one checkpoint under history-depth and prior-assignment interventions, respectively. On the full test set, uniform weighting assigns equal mass to valid priors, whereas permutation preserves the learned values but breaks their prior identities.

On the fixed cohort, increasing the visible history from one to five priors improves both recall and F1. Holding patient composition fixed attributes this trend more directly to deeper visual history. On the full test set, learned assignment produces higher recall and F1 than uniform weighting; relative to permutation, it matches precision and improves the other reported metrics. Together, these controls support the benefits of deeper history and prior-specific assignment; complete results are reported in Supplementary Table~S3.

\paragraph{Natural history availability.}
Table~\ref{tab:natural_depth} compares TALON with the strongest baseline, PriorRG, across natural-history strata; Supplementary Table~S4 provides the complete comparison with MLRG and cohort sizes. TALON exceeds both longitudinal baselines on precision, recall, F1, and RadGraph-F1 in every subset with history. Its F1 advantage generally widens as more prior examinations become available and is largest in the deepest-history subset, complementing the fixed-cohort intervention. The heuristic temporal-change subset shows the same advantage but serves as a transparent stress test rather than temporal-relation ground truth.

\paragraph{Qualitative analysis.}
Case~1 (Figure~\ref{fig:case}) contrasts PriorRG's acute pneumonia overcall with TALON's preservation of chronic hemidiaphragm elevation and atelectasis/scarring. The effective stability coefficients $\alpha_n^{\mathrm{sim}}(1-g_n)$ sum to 0.943, whereas the effective change coefficients $\alpha_n^{\mathrm{chg}}g_n$ sum to 0.056. Thus, the low gates suppress the change stream despite its attention peak at Hist~2, allowing long-horizon stability evidence to prevail.

In the short-interval change case, TALON captures new left-lower-lobe consolidation while retaining chronic right-sided postoperative findings; PriorRG disperses the acute evidence and adds unsupported tube removal. The largest effective change coefficient is assigned to Hist~1 ($\Delta t=21$ days; $w_1^{\mathrm{chg}}=0.162$), while stability evidence remains distributed across deeper priors. This behavior is consistent with using recent history for interval-change localization without discarding chronic context. Complete per-prior attention, gate, and effective routing coefficients are reported in Supplementary Table~S6. These values are descriptive internal outputs rather than causal attributions, and TALON's unsupported small left effusion remains a residual error.

\section{Conclusion}

We presented TALON, a longitudinal RRG framework for variable-length patient histories. Its Dual-Channel Temporal Fusion Module scores each prior for stability and interval change before gated aggregation, preserving persistent findings while emphasizing clinically meaningful changes. On MIMIC-CXR, TALON achieves leading clinical efficacy and RadGraph-F1, with gains that generally grow with history depth. These results support prior-wise, role-dependent selection. TALON currently uses one radiograph per study and five priors; future work will extend it to multi-view inputs and longer trajectories.

\bibliography{aaai2027}

\end{document}